%% file: 0_main.tex
\pdfoutput=1

\documentclass[letterpaper, 10 pt, conference]{ieeeconf}  % Comment this line out if you need a4paper

\IEEEoverridecommandlockouts                              % This command is only needed if 
\usepackage[backend=bibtex,
hyperref=false,
url=false,
isbn=false,
doi=false,
backref=false,
style=numeric-comp,
sortcites,
block=none]{biblatex}

\renewcommand{\bibfont}{\small}
\usepackage{graphicx}
\usepackage{amsmath} % assumes amsmath package installed
\usepackage{amssymb}  % assumes amsmath package installed
\usepackage{textcomp}
\usepackage{wrapfig}
\usepackage{color}
\usepackage{xspace}
\usepackage{overpic}
\usepackage{booktabs}
\usepackage{algorithm2e}
\usepackage{hyperref}
\usepackage{comment}
\usepackage{multirow}
\usepackage{adjustbox}
\usepackage{array}
\usepackage{booktabs}
\usepackage{colortbl}
\usepackage{float,wrapfig}
\usepackage{hhline}
\usepackage{multirow}
\usepackage{subcaption} % issues a warning with CVPR/ICCV format
\usepackage{changepage}
\usepackage{extramarks}
\usepackage{fancyhdr}
\usepackage{lastpage}
\usepackage{setspace}
\usepackage{soul}
\usepackage{xspace}

\definecolor{blue}{rgb}{0,0,1}
\definecolor{red}{rgb}{1,0,0}
\definecolor{green}{rgb}{0,.5,0}
\definecolor{orange}{rgb}{0.75, 0.4, 0}
\definecolor{teal}{rgb}{0.0, 0.4, 0.4}
\definecolor{purple}{rgb}{0.65,0,0.65}

\newcommand{\rh}[1]{{\color{black}\textbf{}#1}\normalfont}

\title{\LARGE \bf
NIFT: Neural Interaction Field and Template for Object Manipulation
}

\author{\authorblockA{Zeyu Huang$^{1}$, Juzhan Xu$^{1}$, Sisi Dai$^{2}$, Kai Xu$^{2}$, Hao Zhang$^{3}$, Hui Huang$^{1}$, Ruizhen Hu$^{1,*}$ \\
$^1$Shenzhen University $\quad$ $^2$National University of Defense Technology $\quad$ $^3$Simon Fraser University\\
$^*$Corresponding Author}}%
\begin{document}

% \maketitle

\twocolumn[{%
	\renewcommand\twocolumn[1][]{#1}%
	\maketitle
	\input{figures/teaser}
}]

\thispagestyle{empty}
\pagestyle{empty}

%%%%%%%%%%%%%%%%%%%%%%%%%%%%%%%%%%%%%%%%%%%%%%%%%%%%%%%%%%%%%%%%%%%%%%%%%%%%%%%%
\begin{abstract}
We introduce NIFT, {\em Neural Interaction Field and Template\/}, a descriptive and robust interaction representation of object manipulations to facilitate imitation learning.
Given a few object manipulation demos, NIFT guides the generation of the interaction imitation for a new object instance %, from the same category as in the demos, 
by matching the \emph{Neural Interaction Template} (NIT) extracted from the demos %to 
in the target \emph{Neural Interaction Field} (NIF) defined for the new object. 
Specifically, the NIF is a neural field that encodes the relationship between each spatial point and a given object, where the relative position is defined by a {\em spherical distance function\/} rather than occupancies or signed distances, which are commonly adopted by conventional neural fields but less informative.
For a given demo interaction, the corresponding NIT is defined by a set of spatial points sampled in the demo NIF with associated neural features. 
To better capture the interaction, the points are sampled on the Interaction Bisector Surface (IBS), which consists of points that are equidistant to the two interacting objects and has been used extensively for interaction representation.
With both point selection and pointwise features defined for better interaction encoding, NIT effectively guides the feature matching in the NIFs of the new object instances such that the relative poses are optimized % optimize the object poses %so as to of the anchor object, i.e., the robot gripper for grasping a mug and the rack for hanging a mug, to 
to realize the manipulation while imitating the demo interactions.
Experiments show that our NIFT solution outperforms state-of-the-art imitation learning methods for object manipulation and generalizes 
%task when given a few demonstrations and generalizes 
better to objects from new categories. 
\end{abstract}

\input{1_introduction}
\input{2_related}
\input{3_method}

\input{4_experiment}

\input{5_conclusion}

\input{6_acknowledge}

\addtolength{\textheight}{-3cm}   % This command serves to balance the column lengths

%%%%%%%%%%%%%%%%%%%%%%%%%%%%%%%%%%%%%%%%%%%%%%%%%%%%%%%%%%%%%%%%%%%%%%%%%%%%%%%%

%%%%%%%%%%%%%%%%%%%%%%%%%%%%%%%%%%%%%%%%%%%%%%%%%%%%%%%%%%%%%%%%%%%%%%%%%%%%%%%%
%\section*{APPENDIX}


%%%%%%%%%%%%%%%%%%%%%%%%%%%%%%%%%%%%%%%%%%%%%%%%%%%%%%%%%%%%%%%%%%%%%%%%%%%%%%%%

%References are important to the reader; therefore, each citation must be complete and correct. If at all possible, references should be commonly available publications.

%\bibliographystyle{IEEEtran}
%\bibliography{IEEEabrv,IEEEexample}
\renewcommand*{\bibfont}{\footnotesize}
\begin{flushright}
	\printbibliography %for biblatex
\end{flushright}

\end{document}

%% file: figures/teaser.tex
%\begin{figure*}
\begin{center}
	\centering
	\captionsetup{type=figure}
	\includegraphics[width= \linewidth]{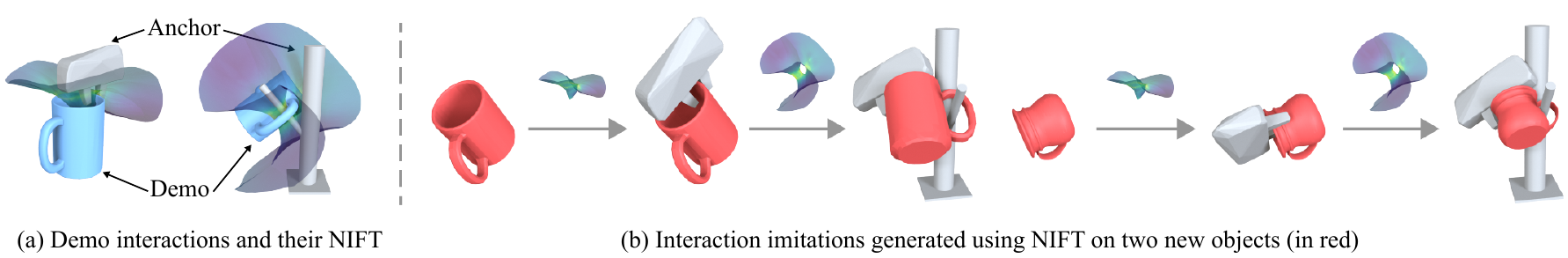}
	%\caption
	\captionof{figure}
%	{Given a few demo interactions of an object manipulation task, we introduce the Neural Interactions Fields and Templates (NIFT) to represent the interaction and guide the interaction imitation for manipulation on new object instances given in arbitrary poses. A Neural Interactions Field (NIF) is a neural field defined around an object to characterize the relationship between any spatial point and the object, and the Neural Interaction Template (NIT) is composed of a set of points sampled in the field and thus associated with the neural features to encode the interactions shown in the demonstrations and is used to guide the matching in the NIF of new object instances.}
	{We introduce NIFT, \textit{Neural Interaction Field and Template}, to represent object interactions for imitation learning.
	Given a few demo interactions (a) for an object manipulation task, NIFT guides the generation of interaction imitations to manipulate new object instances (in red) provided in arbitrary poses (b). %Specifically, a \zh{Neural Interaction Field (NIF)} is a neural field defined around an object to characterize the relationship between any spatial point and the object, while its associated Neural Interaction Template (NIT) is composed of points sampled \rz{from} the NIF with associated neural features to encode the interactions shown in the demos. The NIT is further employed to guide the \zh{feature} matching in the NIF of new object instances.
 }
	\label{fig:teaser}
%\end{figure*}
\end{center}

%% file: 1_introduction.tex
%%%%%%%%%%%%%%%%%%%%%%%%%%%%%%%%%%%%%%%%%%%%%%%%%%%%%%%%%%%%%%%%%%%%%%%%%%%%%%%%
\section{INTRODUCTION}

Teaching a robot with few-shot demonstrations has been a long-standing goal in robotics~\cite{alissandrakis2005approach, dang2010robot, suay2012practical, li2015teaching, huang2016modular, koenig2017robot, argall2009survey}.
Ideally, robots should be able to learn from a few demonstrations of a manipulation task and then generalize to new
instances of target objects. 
Especially for object manipulation tasks such as using a robot gripper to pick up a mug or hanging it on a rack (see Figure~\ref{fig:teaser}), the gripper and the rack are always fixed, which can be considered as the anchor objects, and the goal is usually to perform similar manipulation or interaction to new object instances to imitate the interactions shown in the demonstrations.
If we refer to the source object in the demonstrations as a demo object and the new object instance as the target, our goal then is to generate the interaction between the anchor object and the target so that it is analogous to the interaction between the demo object and the anchor object as shown in the demonstrations.

To optimize the pose of the anchor object with respect to a new target object to mimic the demo interaction,
we need to address fundamental questions on how to represent and optimize 3D interactions, and 
how to measure similarity between the target and the source, or demo, interactions.
%
%Of the most importance 
The most important task is to find the right representation, which should encode rich information of %be sensitive to and characteristic of 
object-object interactions, rather than the individual 3D objects as well as robust against %shape variations
variation in shapes of the interacted objects. 

\input{figures/overview}

Recent work on neural descriptor fields (NDF)~\cite{simeonov2021neural} aims to solve this problem with a neural interaction representation and optimization via feature matching in the neural fields. 
NDF characterizes object interactions by sampling a fixed set of query points, called a Basis Point Set (BPS)~\cite{prokudin2019efficient}, around the anchor object.
However, BPS has been shown to provide an efficient and compact means to encode features of the anchor object alone, and not the interaction between the anchor and other objects.
Moreover, their point-wise features are learned with a network that predicts the occupancy of each point relative to the given object; no spatial relationship between the point and the
object is encoded.
%the pointwise feature used in~\cite{simeonov2021neural}  relies on the network that predicts the occupancy of each point relative to the given object,  and thus spatial relation between the point and the object has not been fully exploited. 

Our key contribution is the introduction of {\em neural interaction field and template (NIFT) \/} to provide an informative interaction representation and similarity measure 
that fulfills the criteria mentioned above. Our representation is designed to effectively guide the feature matching for interaction imitation adaptive to the new geometry of a target object. 
%
%NIFT is inspired by the work of Zhao et al.~\cite{zhao2016relationship}, where a handcrafted scene template is proposed to guide the generation of scene variations. 
%To build the neural interaction template (NIT), we first
%

To encode the open space around the object, we first build the {\em neural interaction field (NIF)}. % around the object.
For each spatial point, rays are cast from this point in all directions, and the distances to the object in all directions form a normalized spherical function. 
Then the spherical harmonics expansion of this spherical function is computed to get the rotation-invariant space coverage feature (SCF)~\cite{zhao2016relationship} of this point relative to the object. 
However, the computation of the SCF is time-consuming and non-differentiable, so instead of using SCF directly, we %follow the work of NDF~\cite{simeonov2021neural} to 
train a neural network to predict the pointwise SCF and then use the vector of concatenated activations as the neural feature per point, which constitutes the NIF of a given object. %which we denoted as the {\em neural interaction field (NIF)\/}. 
For the given demo interaction, we sample a set of Interaction Bisector
Surface (IBS)~\cite{zhao2014IBS} points that are equidistant to two interacting objects.
IBS has been shown to be an informative spatial descriptor of object-object interactions, while robust against shape variations,
and thus we denote this set of IBS points associated with the NIF feature as our {\em neural interaction template (NIT)\/}. 
Then given a new target object, our goal is to find the optimal pose of the anchor object together with the NIT in the target NIF with matched features.

We conduct experiments on three pick-and-place tasks to show that NIFT-guided interaction imitation outperforms the state-of-the-art methods by at least a 10\% overall success rate boost. 
With the more descriptive and robust interaction representation, our method can also generalize well to out-of-distribution objects.
Ablation studies are also presented to show the importance of both our point selection and neural feature design.
%Other than experiments in the simulated environment, 
Apart from experiments in a simulated environment, we also validate our method on a real robot.

%% file: figures/overview.tex
\begin{figure*}[!t]
    \centering
    \includegraphics[width= \textwidth]{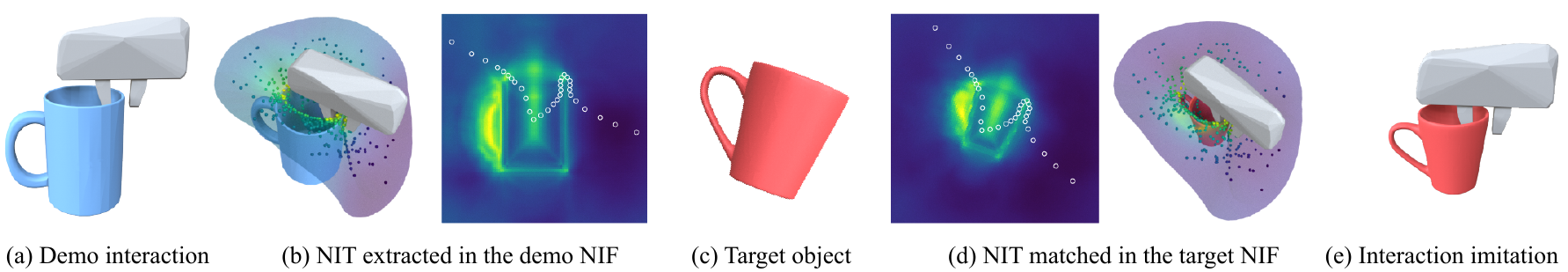}
\caption{\rh{ Overview of our interaction imitation method. (a) Given a demo interaction with a robot gripper picking a blue mug, (b) we first represent the interaction with the %neural interaction template (NIT)
NIT, which consists of points sampled on the IBS associated with features defined in %a neural interaction field (NIF) 
the NIF of the blue mug. (c) %For a target mug in a random pose to generate an analogical interaction,
To generate an analogical interaction for a target red mug in a random pose, (d) we perform the interaction imitation by optimizing the global transformation of the NIT to find its best match in the NIF of the red mug. (d) Finally, the gripper is transformed based on the optimized pose of NIT to form the imitated interaction.}}% with the new object.}
\label{fig:overview}
\end{figure*}

%% file: 2_related.tex
\section{RELATED WORK}

Our work is related to geometric representations of interactions and imitation learning for manipulation.
In this section, we cover prior works most relevant to our method.

\textbf{Interaction representation.}
Interaction representations have been extensively studied to encode spatial relations between two or more objects. 
Relative vector used in most of works for scene generation~\cite{fisher2011characterizing} is relatively simple and thus usually need to be incorporated with other properties together to be able to characterize the %accurate spatial relationship between two objects. 
spatial relationship between two objects accurately.
On the contrary, the %Interaction Bisector Surface (IBS) introduced in Zhao et al.
IBS~\cite{zhao2014IBS} provides more detailed and informative interaction representation with the geometric and topological features extracted from the spatial boundary between two objects, which have been used for scene completion and synthesis~\cite{zhao2016relationship, zhao2019localization}.
The corresponding regions on the interacting objects, denoted as Interaction Region (IR), are further explored for functionality analysis of 3D shapes~\cite{hu2015interaction, hu2016learning}. 
Pirk et al.~\cite{pirk2017understanding} further introduce interaction landscapes to build a spatial and temporal representation of interactions that considers dynamic changes of the interaction between two objects.
However, all of these previous works only focus on the interaction analysis between two objects without considering the possibility of the transition from one representation to the other to make it applicable for  interaction imitation.

\textbf{Imitation learning for manipulation.}
%Ideally, a robot can learn from only a few demonstrations and generalize well to new object instances, and thus 
Imitation learning for manipulation has been extensively studied in robotics.
Based on different assumptions on the difference between the demo and target objects, the design of the method %will focus
focuses on different aspects and thus have different levels of generality. 
For example, pose estimation~\cite{yoon2003real, zhu2014single, schulman2016learning} is the key for known objects, while primitive-based template-matching~\cite{ harada2013probabilistic, rodriguez2018transferring, thompson2021shape} %miller2003automatic
can be more robust to shape and pose changes. 
%To direct learn the manipulation policy, large amounts of demonstrations are usually required
To learn the manipulation policy directly, usually a large number of demonstrations are required~\cite{berscheid2020self, gualtieri2018pick, song2020grasping, she2022learning}.
%For example, Yu et al.~\cite{yu2018one} propose one-shot learning from a video of a human conducting a variety of previous tasks to build up prior knowledge through meta-learning. Combining this prior knowledge and only a single video demonstration from a human, the robot can perform the task that the human demonstrated. 
%The interaction analogy problem studied in this work is related to the one-shot imitation learning as only a single demonstration as input for one-shot learning, while the main difference is that one-shot imitation learning in robotics usually involves learning policy for more complex behaviour while we focus more on unit action like interaction snapshot.
There are also many recent works that use category-level keypoints as an object representation for transferable robotic manipulation~\cite{gao2021kpam, florence2018dense, sundaresan2020learning}. 
The most related work in this direction is NDF~\cite{simeonov2021neural}, following the idea of using neural shape representation~\cite{chen2019learning, niemeyer2019occupancy, park2019deepsdf, peng2020convolutional, rebain2021deep}.
NDF divides the whole object manipulation task into two stages: interaction synthesis and motion planning, and in the few-shot imitation setting, the interaction synthesis task here is essentially to perform interaction imitation. 
In this work, we focus on solving the interaction imitation problem with the same pipeline as in NDF~\cite{simeonov2021neural} but using the proposed NIFT to represent the demo interactions to guide and improve the performance of few-shot imitation learning of 3D interaction.
%Thus our method can be applied in this setting, with the key difference being the interaction representation and how the equidistant constraint of our representation enables deformations of the anchor object to improve the performance of one-shot learning for generative modeling of 3D interactions.

%% file: 3_method.tex
\section{METHOD}
Given a demo interaction $(O_a, O_s)$, where $O_a$ is the anchor object and $O_s$ is the source object that is to be replaced with a target object $O_t$ given in a random pose, our goal is to optimize the global pose $T$ of the anchor object $O_a$ w.r.t $O_t$ such that their interaction imitates the demo interaction.

An overview of our method is illustrated in Figure~\ref{fig:overview}. %, which shows an example of deformable interaction analogy on hand grasping.
Given the demo interaction shown on the left, we first represent the interaction by sampling the IBS points  $\mathcal{X}_{sa}$ %(on the surface shown in blue) 
on the transparent surface in the NIF of the demo object to form the NIT. 
The global pose of $\mathcal{X}_{sa}$ is then optimized relative to the target object $O_t$ to form an interaction imitation, based on the feature matching in the target NIF.
So the key idea of our method is to use the NIT to represent the interaction between  $O_a$  and $O_s$,  and further guide the global transformation of the $O_a$  towards $O_t$ based on the feature matching in the target NIF to achieve the interaction imitation.

\subsection{Neural Interaction Field (NIF)}
%For each spatial point around a given object, we first compute the SCF~\cite{zhao2016relationship} to provide effectively quantify the relationship between the point and the object, which encodes the geometry of the open space around objects in the frequency domain using spherical harmonics. As shown in Figure~\ref{fig:SCF}, for each spatial point $x$, rays are cast from $x$ in all directions to compute a normalized spherical function $F_x$:
\input{figures/net}

\rh{The definition of NIF around a given object is inspired by NDF~\cite{simeonov2021neural}. As shown in Figure~\ref{fig:NIF}, % (bottom), 
we train a network to predict the spatial feature of a point $x$ relative to the object $O$. However, instead of predicting occupancy of each query point as in NDF~\cite{simeonov2021neural}, we use more informative SCF~\cite{zhao2016relationship} to effectively quantify the relationship between the point and the object. % to form our NIF. 
The %informative 
concatenated activations of the network decoder $\Phi$ is then used as the point descriptor in the NIF:
\begin{equation}
	f(x|O) = \mathop{\bigoplus }_{i=1}^{L} \Phi^{i}(x, \epsilon(O) ),
\end{equation}
where $\epsilon$ is a point cloud encoder, $\Phi^{i}$ is the activation of the $i$-th layer of the decoder, $L$ is the total number of layers of the decoder, and $\bigoplus $ denotes concatenation. 
Note that SCF is SO(3)-invariant, and we use the SO(3)-equivariant network architecture proposed in~\cite{deng2021vector} %for $\Phi$ 
as in NDF~\cite{simeonov2021neural} to achieve SO(3)-equivariance such that the descriptor of a point $x$ remains constant when its relative position to the object $O$ is fixed, regardless of the change of their global configuration. % (3D rotation).

%to further achieve SO(3)-equivariance such that the descriptor of a point $x$ remains constant when its relative position to the object $O$ is fixed, regardless of the change of their global configuration (3D rotation), we use the SO(3)-equivariant architecture proposed in~\cite{deng2021vector} for $\Phi$ as in~\cite{simeonov2021neural}.  %Translation equivariance is conveniently implemented by subtracting the point cloud mean from both the point cloud and the coordinate. We thus re-define f(x|P) as:
%The SE(3)-equivariance is achieved by translating both the query point and input object to the local frame of the object centroid before feature extraction. 
%\input{figures/SCF}

%
%\textbf{SCF computation.}
%More specifically, 
In more details, SCF encodes the geometry of the open space around objects in the frequency domain using spherical harmonics. %As shown in Figure~\ref{fig:NIF} (top), 
For each spatial point $x$, rays are cast from $x$ in all directions to compute a normalized spherical function $F_x$:}
\begin{equation}
	\label{eq:scf}
	F_x{(\theta,\phi)} = \frac{d_{min}+d_{avg}}{d_x{(\theta,\phi)}+d_{avg}} 
\end{equation}
where $d_x(\theta,\phi)$ is the hit distance of the ray along the direction $(\theta,\phi)$, and $d_{min}$ and $d_{avg}$ are the minimum and mean value of all non-infinity distance of $d_x$.
%If a ray does not hit an object, the distance is set to infinity. 
%Then the spherical harmonics expansion~\cite{kazhdan2003rotation} of  $F_x$ is computed: 
Then $F_x$ is decomposed into weighted sum of a group of spherical function basis via spherical harmonics expansion~\cite{kazhdan2003rotation}: 
\begin{equation}
	F_x{(\theta,\phi)} = \sum_{l=0}^{\infty}\sum_{|m| \leq l}{c_{l}^{m}Y_{l}^{m}(\theta,\phi)}
\end{equation}
where $c_{l}^{m}$ are the spherical harmonics coefficient and $Y_{l}^{m}$ are the orthonormalized spherical harmonics at frequency $l$. 
The SCF of point $x$ is finally defined as:
\begin{equation}
	SCF_x = \{c_0, c_1, ..., c_n\}
\end{equation}
where $c_{l} = \lVert \sum_{|m|\leq l}{c_{l}^{m}Y_{l}^{m}} \rVert_2 = \sqrt{\sum_{|m| \leq l}(c_{l}^{m})^2}$ is the power of the function at frequency $l$.% In all our experiments, we set $n=5$.

%As the computation of SCF is rather complicated and time-consumption, it would be inefficient to use SCF directly for interaction representation to guide further optimization. 
%Following the key idea of NDF~\cite{simeonov2021neural}, we introduce a neural interaction field around the object based on SCF.
%In more detail, we train a network $\Phi$ to predict the SCF feature of a spatial point $x$ relative to the object $O$ instead of the occupancy.
%As the encoder is a hierarchical, coarse-to-fine feature extractor that encodes the information required,  the concatenated activation of the encoder is then used as the final point descriptor:
% \begin{equation}
% 	f(x|O) = \mathop{\bigoplus }_{i=1}^{L} \Phi^{i}(x, \epsilon(O) ),
% \end{equation}
% where $\epsilon$ is the point cloud encoder, $\Phi^{i}$ is the activation of the $i$-th layer, $L$ is the total number of layers, and $\bigoplus $ denotes the concatenation operator.
% \input{figures/net}

%\input{figures/SCF_example}

%The field is defined as a function $f$ that maps a 3D coordinate of a point $x$ in the local frame of the object to a spatial descriptor $z = f(x|O) \in \mathbb{R}^{n}$ encoding the spatial relationship of point $x$ w.r.t. $O$.

\input{figures/opt_diff_point}

Figure~\ref{fig:opt_diff_point} shows some comparisons of different pointwise features, including SCF~\cite{zhao2016relationship} predicted by the network, NDF~\cite{simeonov2021neural} based on occupancy, and our NIF based on SCF. 
For a spatial point around the demo object, we compute the L1 feature differences of this point to all the points around the target object and show a slice of the difference map, where blue indicates smaller difference and red indicates larger difference.
%Blue regions in the difference map are close to the spatial point in the feature space, while red regions are far from the spatial point. 
%Note that the network we trained to predict SCF has high accuracy and the corresponding difference maps are quite similar.
We can see that compared to NDF and the predicted SCF,  NIF is more distinctive in corresponding to the similar spatial regions of the different objects.
\rh{Note that as SCF encodes relative distance information in a normalized scale according to Equation~\ref{eq:scf}, our NIF is also robust to the scale differences between the demo and target objects.}
%\rz{Note that we assume that the source objects and test objects are in similar sizes in our experiments. In cases where the source and target objects are of drastically different sizes, we can obtain similar results for regions close to the object surface since NIF encodes relative distance information in a normalized scale according to Equation~\ref{eq:scf}.}

%\subsection{Neural interaction template}

\subsection{Neural Interaction Template (NIT)}

As the goal is to transfer the anchor object $O_a$ to the target object $O_t$ to imitate the interaction between $O_a$  and $O_s$, we would like the interaction representation %captures the feature of the most important region 
to capture the features of the most important regions related to the interaction.
%Moreover, to be able to deal with the target object given in arbitrary poses, the features should be SE(3)-invariant.
Thus, we opt to sample a set of IBS points $\mathcal{X}_{sa}$ in the demo NIF of $O_s$ %with accosiated pointwise features $\{f(x|O_s)\}$  
to form the NIT representing the demo interaction between $O_a$  and $O_s$.

Given a demo interaction with a pair of objects $(O_a, O_s)$, IBS~\cite{zhao2014IBS} is defined as a set of points that are equidistant to both the objects.
To compute IBS, we first sample a set of points on the surfaces or point clouds of the two interacting objects uniformly and compute the Voronoi diagram for all those samples. 
The IBS is a subset of the Voronoi diagram which lies in between the two objects.
Since IBS extends infinitely, we truncate it where it intersects with the bounding sphere of the two objects, and then 
%sample a set of query points $\mathcal{X}_{sa}$, on the IBS as in~\cite{zhao2014IBS}.%to be the query point $\mathcal{X}_{sa}$.
use an importance-based sampling scheme to sample a set of points on the IBS to be the query point cloud $\mathcal{X}_{sa}$ as in~\cite{zhao2014IBS}. Intuitively, regions which are closer to the interacting objects are more important for characterizing the interaction and thus more points are sampled from there to give a better abstraction of the IBS.

\subsection{Pose optimization for interaction imitation}
\label{sec:pose}

%As the source object $O_s$ gets replaced with a target object $O_t$ while the anchor object $O_a$ remains fixed during imitation learning, for each point $x \in \mathcal{X}_{sa}$, we compute its feature conditioned on the source object  $f(x|O_s)$ to make sure it encodes information of $O_s$ when being transferred together with $O_a$ to $O_t$.

The goal of interaction imitation is to find an optimal pose of $O_a$ so that the spatial relationship between $O_a$  and $O_t$  resembles the one between $O_a$ and $O_s$.
With the interaction between $O_s$ and $O_a$  being represented by IBS points $\mathcal{X}_{sa}$ and the associated demo NIF features $\{ f(x|O_s)\}$, the goal now becomes to find an optimal pose of $\mathcal{X}_{sa}$ such that the point-wise features conditioned on target NIF $\{ f(x|O_t)\}$ are as close to $\{ f(x|O_s)\}$ as possible. 
This is because the relative pose between $\mathcal{X}_{sa}$ and $O_a$  is fixed, and the global pose of $O_a$ is uniquely determined by that of $\mathcal{X}_{sa}$.
Thus, the optimization is formulated as:
\begin{equation}
	\label{eq:obj}
	\bar{T} = \mathop{\arg\min}_{T}\\ \mathop{\sum}_{x \in \mathcal{X}_{sa}}\lVert f(x|O_s) - f( Tx|O_t ) \rVert_1.
\end{equation}
% where the L1 distance between the descriptors is used.

To optimize the global pose of $\mathcal{X}_{sa}$ relative to $O_t$, we first move the centroid of $\mathcal{X}_{sa}$ to the centroid of $O_t$,
%We randomly initialize the pose $T = (R, t)$ while confining each translation component into $[0, 0.1]$. We then optimize the rotation $R$ and translation $t$ to minimize the objective function in Eq. (\ref{eq:obj}) using an iterative optimization solver, in particular, ADAM~\cite{kingma2014adam}.
and then randomly sample a translation $t$ near the origin and a rotation $R$ from the Haar distribution to compose the initial pose $T = (R, t)$. 
We then optimize the rotation $R$ and translation $t$ to minimize the objective function in Eq. (\ref{eq:obj}) using an iterative optimization solver, in particular, ADAM~\cite{kingma2014adam}. 

\input{figures/opt_diff_feature}

Figure~\ref{fig:opt_diff_feature} shows the comparison of the optimization process using either IBS points or BPS points but with the same pointwise feature NIF. 
We color-code the points to indicate the point-wise feature distances. 
We can see that in both cases the gripper (anchor object $O_a$) gradually moves closer to the mug (target object $O_t$), guided by the movement of the query points $\mathcal{X}_{sa}$, and finally forms an interaction analogy. Note how the distances of the points are minimized during the optimization, while IBS leads to more accurate interaction without penetration.

\subsection{Few-shot imitation learning}
When given more than one demonstration, we incorporate the information extracted from all demo interactions to form the NIT. We first align the IBS points from all demo interactions into a common frame determined by the anchor object, and then perform density-based resampling to keep the points most relevant to the interaction on this anchor. In more detail, we compute a normalized weight for each point based on the average distance to its $k$ nearest neighbors to resample the points. We set $k$ equal to the number of demonstrations in our experiments. We take the resampled points as the query points of NIT to compute the NIF features for each demo object, and the average NIF feature of each point is used as the final pointwise feature of NIT. Once we have an NIT, we can perform imitation learning by performing the optimization explained in Sec. \ref{sec:pose}.

%% file: figures/net.tex
\begin{figure}[!t]
	\centering
	\includegraphics[width=\linewidth]{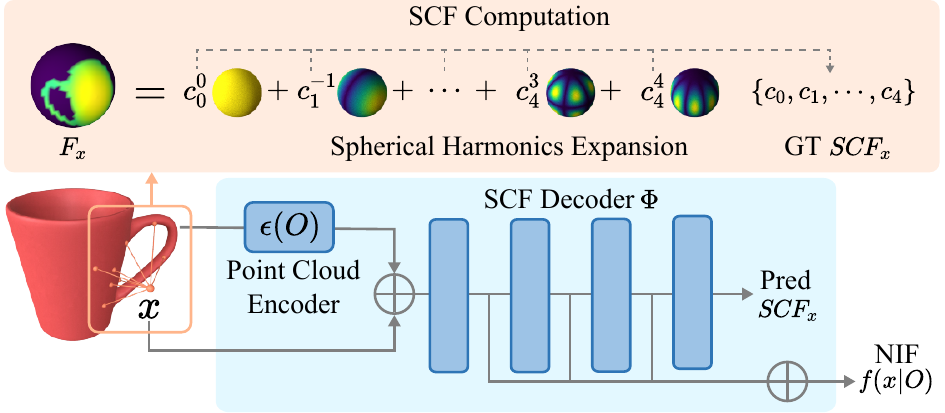}
	\caption{
		\rh{NIF computation. For each spatial point around the object, we predict its SCF using a neural network and concatenate the activation of the network decoder as its feature in the neural interaction field. }}
	\label{fig:NIF}
\end{figure}

%% file: figures/opt_diff_point.tex
\begin{figure}[!t]
	\centering
	\includegraphics[width=\linewidth]{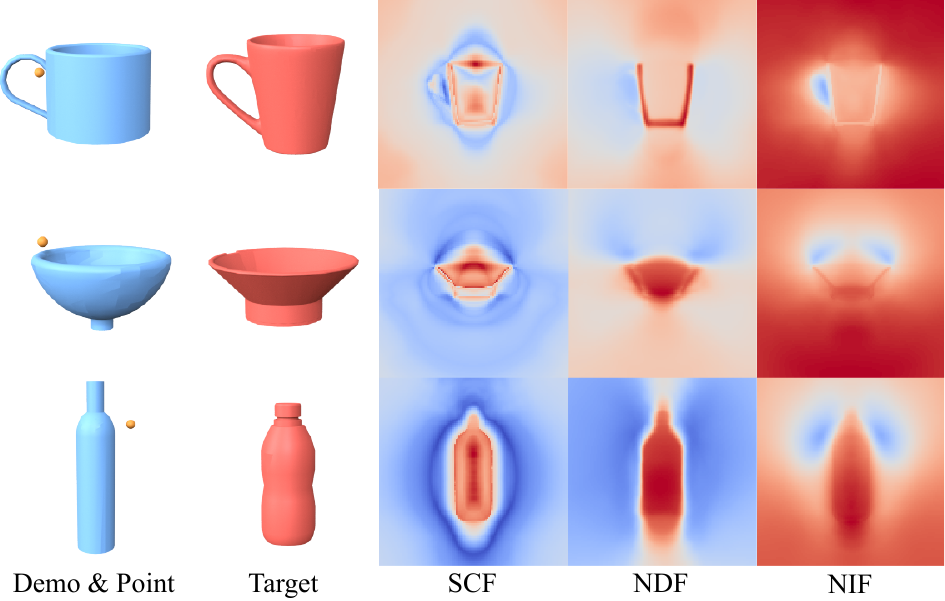}
	\caption{
		% 3维模型上一个点，SCF-BPS优化的对应点以及特征error图，SCF-IBS优化的对应点以及特征error图
		Comparison of different pointwise features. In each row, for a spatial point located around the source object, we compute the feature differences of this point to all the points around the target object and show the difference map.   }
	\label{fig:opt_diff_point}
\end{figure}

%% file: figures/opt_diff_feature.tex
\begin{figure}[!t]
	\centering
	\includegraphics[width=\linewidth]{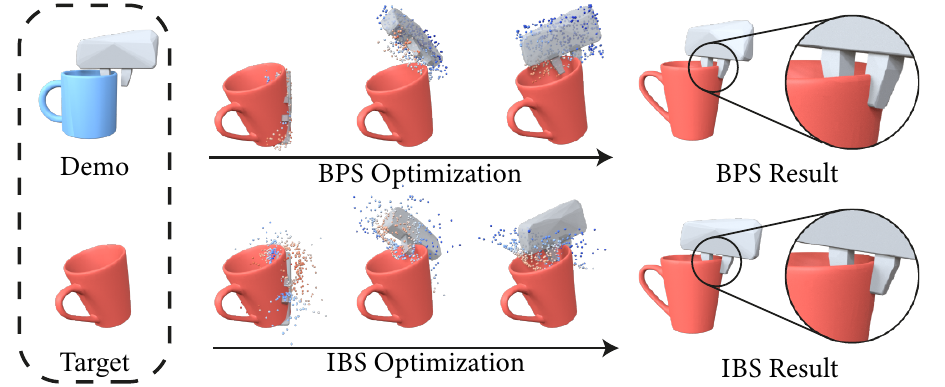}
	\caption{
		% 3维模型上一个点，OCC-IBS优化的对应点以及特征error图，SCF-IBS优化的对应点以及特征error图
		Comparison of the optimization process using either IBS points or BPS points, both associated with NIF features.}% while NIF is used for pointwise features in both cases.}
	\label{fig:opt_diff_feature}
\end{figure}

%% file: 4_experiment.tex
\section{EXPERIMENT}

%The work of ~\cite{simeonov2021neural}

%Since we use the same point-wise feature (NDF) and similar optimization as in~\cite{simeonov2021neural}, we mainly compare baselines with different ways of query point set selection.

% How well do SCF\_NDF enable manipulation of unseen objects in unseen poses?
% What impact does the parameterization of NDFs have on its performance?
% Can NDFs transfer to a real robot?

\input{figures/tab_rigid}

\subsection{Experiment setup}
% general task
\textbf{Task setup.} To evaluate our method, % and compare it to previous methods, 
we conduct the same pick-and-place experiments introduced in NDF~\cite{simeonov2021neural}: 1) Grasp the rim of a \emph{mug} and hang its handle to a rack; 2) Grasp the side of a \emph{bowl} and place it upright on a shelf; and 3) Grasp the neck of a \emph{bottle} and place it upright on a shelf.
%\begin{itemize}
%	\item Grasp the rim of a \emph{mug} and hang its handle to a rack.
%	\item Grasp the side of a \emph{bowl} and place it upright on a shelf.
%	\item Grasp the neck of a \emph{bottle} and place it upright on a shelf.
%\end{itemize} 
Given a few demonstrations of the same task with objects initialized in the upright pose, the robot is asked to manipulate unseen objects in similar ways.

% environment
\textbf{Environment setup.} %Quantitative evaluation is conducted in the simulated environment. 
Following NDF~\cite{simeonov2021neural}, we create the simulated environment in PyBullet~\cite{coumans2021} with a two-finger gripper on a table and a RGB-D camera at each corner to obtain the fused point clouds for objects. It is assumed that the object is segmented from the background and the environment remains fixed between demo and test time. 
%A subset of shapes is sampled from ShapeNet~\cite{chang2015shapenet} to train the networks and test in simulated experiments.

% experiment setting
\textbf{Experiment setup.} For each task in the simulated environment, we conduct two groups of experiments with different initial pose settings. For the easy setting, we initialize the test objects with a random upright pose like demonstration. For the hard setting, all objects are initialized in arbitrary SE(3) poses upon the table. We also apply random uniform scaling to all test objects. In each experiment, we provide 10 upright demonstrations and 100 unseen object instances.

% metrics
\textbf{Evaluation metrics.} 
Quantitative evaluation is conducted in the simulated environment, and the results are evaluated by the success rates for grasping, placing, and overall processing. A successful grasping means the test object is stably grasped by the arm after disabling the physical collision between the test object and the other objects in the environment. Similarly, a success of placing needs the object to be stably dropped on the rack or shelf after setting the test object to an optimized pose. Note that we do not require the arm to make a successful motion planning from the grasped pose to the pre-place pose for the success of the overall process. \rh{As in NDF~\cite{simeonov2021neural}, we use off-the-shelf inverse kinematics and motion planning algorithms to execute the final pick-and-place task based on the predicted relative poses, and the collisions between the gripper and the object are ignored before reaching the target pose.}

%Note that we also attempted to compare to other registration baselines including FilterReg~\cite{gao2019filterreg}, Gaussian mixture (GMM) trees~\cite{eckart2018fast}, and point contrast learning~\cite{choy2019fully,xie2020pointcontrast} to find keypoints on the target object corresponding to the demo object as in~\cite{florence2018dense}. However, the CPD baseline was found to be more robust. In particular, point contrast learning appeared unable to learn useful features from our highly similar training data of the same category.}

% training details
\textbf{Training details.} To train %the implicit networks, 
networks in our work and baselines, we use the simulation environment to create a dataset with 100,000 point clouds of randomly scaled and posed objects for each object category and train the network across all categories. We sample spatial points uniformly in a box 1.5 times larger than the object's bounding box and compute the SCF and occupancy as ground truth. We use L1 loss for the SCF network and binary cross entropy loss for the occupancy network. The networks have the same architecture.
We train all the networks with gradient descent using Adam optimizer~\cite{kingma2014adam} with a learning rate 1e-4 for 50 epochs. We use 10\% of the dataset as the validation set.% and use the model weights with minimum validation loss. 
%The accuracy for the occupancy prediction achieves 92\%, and the mean $R^2$ score for the SCF regression achieves 86\%.
The occupancy network achieves 92\% in accuracy and the SCF network achieves 86\% in mean $R^2$ score.

% optimization details
\textbf{Optimization details.} To minimize the objective function of few-shot imitation learning, we use the Adam optimizer with learning rate 1e-2 and optimize it for maximum 500 iterations. Following previous work~\cite{simeonov2021neural}, 10 parallel optimizations with different initialization are conducted at the same time, and the optimized pose with the lowest error is taken as the final result.  All results are converged within the max iteration in our experiments.

% baseline 
\textbf{Baselines.} We compare our method to two types of baselines. The first set of baselines are methods with the same optimization pipeline as ours but using different combinations of query points and point-wise features. For query points selection, we compare IBS with BPS used in~\cite{simeonov2021neural}. % the fixed points uniformly sampled in the bounding box of the anchor object used in the previous work~\cite{simeonov2021neural}, which is denoted as BPS. 
For the point-wise features, we compare NIF with NDF used in~\cite{simeonov2021neural} and SCF directly output by the prediction network. 
\rh{The second type of baseline is a more traditional object-matching method following~\cite{rodriguez2018transferring}. More specifically, we take all demo objects as templates and register them with the target object using coherent point drift (CPD)~\cite{myronenko2010point}. The best registration result is then used to transfer the gripper from the demo object to the target to form the final interaction. }

%To conduct the pick-and-place tasks with a two-finger gripper, three object categories are tested, including \emph{Mug}, \emph{Bowl}, and \emph{Bottle}.
%Mugs and bowls share the same pick action of grasping by the rim while the bottles are grasped from the side near the top.
%The place action is to hang them on a rack for mugs or to place them upright on a shelf for bowls and bottles.
%For each task, 10 demonstrations are provided, and 100 unseen object instances are given with random initial poses and random uniform rescaling for testing.
%To quantitatively evaluate the results, success rates for grasping, placing, and overall processing are measured.

\input{figures/demo}

%We also use their NDF model to compute the pointwise features, which is pre-trained on a dataset of 100,000 objects of mug, bowl and bottle categories at random tabletop poses.

%\paragraph{Baselines.}
%Since we use the same point-wise feature (NDF) and similar optimization as in~\cite{simeonov2021neural}, we mainly compare baselines with different ways of query point set selection. Other than \emph{BPS} used in~\cite{simeonov2021neural} and \emph{IBS} used in our method, which are all point in free space, we also test their corresponding surficial point sets by mapping each point onto the nearest point on the object $\hat{P}$, which can be considered as the interaction region (IR) and has been used for interaction region detection of unseen objects in~\cite{hu2016learning} .
%We denoted those two surficial versions of query points as \emph{BPS-IR} and \emph{IBS-IR}, respectively.

%\input{figures/tab_one-shot2}
%\input{figures/tab_out-domain2}

\subsection{Experiment results}

Table~\ref{tab:rigid} shows the comparison of success rates of \rh{our method to different baselines.
We can see that methods utilizing neural fields consistently outperform the traditional object-matching method, denoted as CPD. CPD requires an initial coarse alignment to avoid local minima, which leads to apparent performance degradation in the arbitrary pose setting.
%obtained using different combinations of query points (BPS and IBS) and pointwise features (NDF, SCF and NIF).
%Note that the combination of  BPS and OCC\_NDF shown in the first row corresponds to the method of~\cite{simeonov2021neural}, and the last row is our NIT with IBS and SCF\_NDF.
When checking different combinations of query points (BPS and IBS) and pointwise features (NDF, SCF and NIF), } we can see that the combination of IBS and NIF yields the best results, especially for the overall success, confirming the superiority of NIT on interaction representation. 

When comparing the baselines with the same pointwise feature but different sets of query points, we see that in most cases IBS produces better results than BPS.
When comparing the baselines with the same set of query points but with different pointwise features, we see that the advantage of NIF over NDF is less dominant. \rh{We believe the more informative features provided by NIF should be used with points located in important regions rather than points all over the space to provide clearer guidance.}

\input{figures/tab_outdomain}

\textbf{Out-of-domain test.}
Table~\ref{tab:outdomain} shows the out-of-domain test of our method comparing to NDF~\cite{simeonov2021neural}. 
In one experiment, the source and target objects are from different categories but can be applied by the same type of grasp or place action. For example, mugs can be grasped by the rim and placed on the shelf as the bowls, so we can use the bowl demonstration but test on a mug. 
In the other experiment, the target objects are sampled from an unseen category. We use vases in ShapeNet~\cite{chang2015shapenet} to imitate the grasping and placing interaction of bottle in the demonstration. We can see that our method generalizes better than \rh{the object-matching baseline CPD and} the work of~\cite{simeonov2021neural} in both cases, and we think it is because our NIT can better represent the characteristics of interaction in the demonstrations instead of the specific object. Occupancy is sensitive to the definition of inside and outside of the global shape, while SCF is a local geometric feature of a spatial point which makes it more suitable to represent interactions between objects.

\textbf{Hyperparameter for SCF computation.}
%The order number $n$ is the main hyperparameter for SCF computation. 
Since SCF is the power of spherical harmonic coefficients, its first few orders are related to the spatial structure around the point, while the latter orders represent more details. 
Table~\ref{tab:scf_order} shows the results when using different SCF orders, $n$. 
We can see that the performance is relatively stable but too few orders ($n=3$) %will make SCF limited in its representation capability. 
limits the representation power of the SCF. On the other hand, SCF with more orders may introduce confusion by its redundant information.
We set $n=5$ in all our experiments.

\textbf{Real world execution.}
%Other than 
Apart from experiments in the simulated environment, we also validate our method on a real robot as shown in Figure~\ref{fig:demo}. 
We use a UR5 arm with a two-finger gripper to execute object manipulation tasks. 
To test the robustness of our method, we use only one KinectV2 RGB-D camera in front of the robot to capture a single-view point cloud of the demonstration and test objects in the local coordinate frame of the robot. % in the base frame of the robot. 
We also retrain the network with only one viewpoint of the generated dataset in the simulation environment.
We record all demonstrations with objects in upright poses and test with unseen objects in the same category. 
For each test,  three demonstrations are used.
With only single-view information and a few demonstrations, NIFT is able to guide the interaction imitation on shapes with various geometries.
Please check the supplementary video for real-world task execution.

\input{figures/tab_scforder}

%% file: figures/tab_rigid.tex
\begin{table*}[!t]%
	\caption{\rh{Comparison of different methods on object manipulation task. CPD uses the demo objects as templates, and other methods use different combinations of query points (BPS and IBS) and pointwise features (NDF, SCF and NIF). Note that the combination of BPS and NDF refers to the method of~\cite{simeonov2021neural}, and our NIT with IBS and NIF is shown in the last row.} }
	\label{tab:rigid}
	%\begin{minipage}{\columnwidth}
		\begin{center}
			\small
			%\footnotesize
\begin{tabular}{ll|ccc|ccc|ccc|ccc} 
\hline
Method &  & \multicolumn{3}{c|}{Mug} & \multicolumn{3}{c|}{Bowl} & \multicolumn{3}{c|}{Bottle} & \multicolumn{3}{c}{MEAN} \\ 
\hline
Point & Feature & Grasp & Place & Overall & Grasp & Place & Overall & Grasp & Place & Overall & Grasp & Place & Overall \\ 
\hline
\multicolumn{14}{l}{Upright Pose} \\
\multicolumn{2}{l|}{CPD} & 0.91 & 0.50 & 0.47 & 0.81 & 0.99 & 0.80 & 0.54 & 1.00 & 0.54 & 0.75 & 0.83 & 0.60 \\ 
\hline
\multirow{3}{*}{BPS} & NDF & 0.97 & 0.92 & 0.89 & 0.96 & 0.83 & 0.81 & 0.85 & 0.98 & 0.85 & 0.93 & 0.91 & 0.85 \\
 & SCF & 0.97 & 0.68 & 0.65 & 0.92 & 0.77 & 0.69 & 0.45 & 0.87 & 0.39 & 0.78 & 0.77 & 0.58 \\
 & NIF & 0.99 & 1.00 & 0.99 & \textbf{0.98} & \textbf{1.00} & \textbf{0.98} & 0.68 & 1.00 & 0.68 & 0.88 & 1.00 & 0.88 \\ 
\hline
\multirow{3}{*}{IBS} & NDF & 0.98 & 0.85 & 0.84 & 0.97 & 1.00 & 0.97 & \textbf{0.93} & 0.99 & \textbf{0.93} & \textbf{0.96} & 0.95 & 0.91 \\
 & SCF & 0.96 & 0.54 & 0.53 & 0.97 & 0.35 & 0.34 & 0.46 & 0.93 & 0.45 & 0.8 & 0.61 & 0.44 \\
 & NIF & \textbf{0.99} & \textbf{1.00} & \textbf{0.99} & 0.96 & 1.00 & 0.96 & 0.90 & \textbf{1.00} & 0.90 & 0.95 & \textbf{1.00} & \textbf{0.95} \\ 
\hline
\multicolumn{14}{l}{Arbitrary Pose} \\
\multicolumn{2}{l|}{CPD} & 0.32 & 0.34 & 0.11 & 0.43 & 1.00 & 0.43 & 0.43 & 0.97 & 0.41 & 0.39 & 0.77 & 0.32 \\ 
\hline
\multirow{3}{*}{BPS} & NDF & 0.68 & 0.63 & 0.47 & \textbf{0.76} & 0.79 & 0.58 & 0.55 & \textbf{1.00} & 0.55 & 0.66 & 0.81 & 0.53 \\
 & SCF & 0.61 & 0.75 & 0.43 & 0.71 & 0.73 & 0.51 & 0.49 & 0.88 & 0.41 & 0.60 & 0.79 & 0.45 \\
 & NIF & 0.65 & 1.00 & 0.65 & 0.67 & 1.00 & 0.67 & 0.46 & 0.97 & 0.45 & 0.59 & 0.99 & 0.59 \\ 
\hline
\multirow{3}{*}{IBS} & NDF & 0.71 & 0.54 & 0.39 & 0.71 & 1.00 & 0.71 & 0.54 & 0.98 & 0.53 & 0.65 & 0.84 & 0.54 \\
 & SCF & 0.65 & 0.68 & 0.44 & 0.79 & 0.44 & 0.31 & 0.44 & 0.89 & 0.36 & 0.60 & 0.67 & 0.37 \\
 & NIF & \textbf{0.74} & \textbf{1.00} & \textbf{0.74} & 0.75 & \textbf{1.00} & \textbf{0.75} & \textbf{0.59} & 0.99 & \textbf{0.59} & \textbf{0.69} & \textbf{1.00} & \textbf{0.69} \\
\hline
\end{tabular}
		\end{center}
	%\end{minipage}
\end{table*}%

%% file: figures/demo.tex
\begin{figure*}
	\centering
	\includegraphics[width= \linewidth]{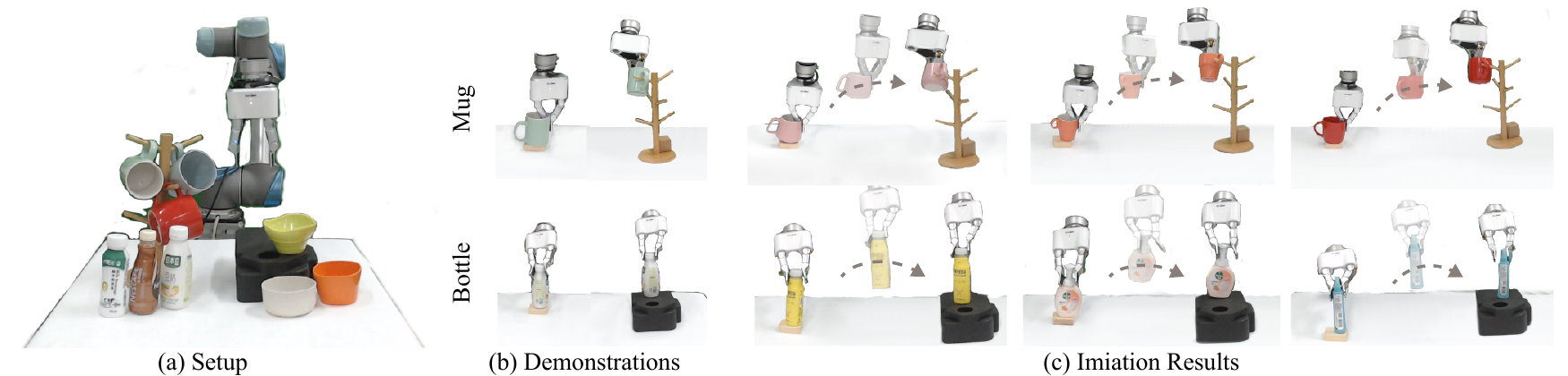}
	\caption{
	Example executions of NIFT for manipulation tasks on real objects. 
	(a) The setup of our experiment with only single RGB-D camera facing the robot. 
	(b) Example demonstrations for the pick-and-place tasks on objects from different categories.
	(c) Example imitation results on new object instances.}
	\label{fig:demo}
\end{figure*}

%% file: figures/tab_outdomain.tex
\begin{table}[!t]%
	\caption{Out-of-domain tests where the source object (Bowl/Bottle) in the demonstration and the testing target object (Mug/Vase) are from different categories.  } 
	\label{tab:outdomain}
	%\begin{minipage}{\columnwidth}
	\begin{center}
		\small
		\footnotesize
\begin{tabular}{l|ccc|ccc} 
\hline
\multirow{2}{*}{Method} & \multicolumn{3}{c|}{Bowl:Mug} & \multicolumn{3}{c}{Bottle:Vase} \\ 
\cline{2-7}
 & Grasp & Place & Overall & Grasp & Place & Overall \\ 
\hline
\multicolumn{7}{l}{\textbf{Upright Pose }} \\
CPD & 0.89 & 0.89 & 0.78 & 0.52 & \textbf{1.00} & 0.50 \\
NDF~ & 1.00 & 0.62 & 0.62 & 0.72 & 0.86 & 0.66 \\
Ours & \textbf{1.00} & \textbf{0.99} & \textbf{0.99} & \textbf{0.92} & 0.99 & \textbf{0.92} \\ 
\hline
\multicolumn{7}{l}{\textbf{Arbitrary Pose}} \\
CPD & 0.56 & 0.94 & 0.53 & 0.32 & 0.89 & 0.30 \\
NDF~ & 0.66 & 0.76 & 0.51 & 0.37 & 0.95 & 0.35 \\
Ours & \textbf{0.73} & \textbf{1.00} & \textbf{0.73} & \textbf{0.56} & \textbf{0.99} & \textbf{0.56} \\
\hline
\end{tabular}
	\end{center}
	%\end{minipage}
\end{table}%

%% file: figures/tab_scforder.tex
\begin{table}[!t]%
	\caption{Test with different SCF orders, $n$. } 
	\label{tab:scf_order}
	%\begin{minipage}{\columnwidth}
	\begin{center}
		\small
		%\footnotesize
\begin{tabular}{c|ccccc} 
	\hline
	SCF Order & 1    & 3    & 5             & 7    & 10    \\ 
	\hline
	Upright Pose        & 0.92 & 0.95 & \textbf{0.95} & 0.94 & 0.93  \\
	Arbitrary Pose       & 0.54 & 0.65 & \textbf{0.69} & 0.66 & 0.62  \\
	\hline
\end{tabular}
	\end{center}
	%\end{minipage}
\end{table}%

%% file: 5_conclusion.tex
\section{CONCLUSIONS}

We introduce neural interaction field and template (NIFT) to provide a descriptive and robust representation of demo interactions as well as guide the imitation learning for object manipulation. 
Experiments show that both point selection and pointwise feature design are essential for the final performance and suggest using the combination of IBS and NIF, leading to our NIT, highly boosts the performance, compared to state-of-the-art methods.

For further investigation, as the NIF we used to compute pointwise features for interaction representation  
is data-driven, although it's already more informative than NDF as shown in our experiments, it may still not generalize well to objects from unseen categories, so it would be interesting to explore other SE(3)-invariant and differentiable feature designs that are more robust to geometric variations. 
Moreover, currently we assume that the anchor objects are fixed, e.g., the two-finger gripper and the hanger, we would also like to further explore ways to perform few-shot imitation learning for object manipulation tasks using %grippers
anchor objects with higher DoF.

%% file: 6_acknowledge.tex
\section{ACKNOWLEDGEMENT}
We thank the anonymous reviewers for their valuable comments and Sai Raj Perla for proofreading the paper. This work was supported in parts by NSFC (U2001206, U21B2023), GD Natural Science Foundation (2021B1515020085), GD Talent Program (2019JC05X328), Shenzhen Science and Technology Program (RCYX20210609103121030), DEGP Innovation Team (2022KCXTD025), and Guangdong Laboratory of Artificial Intelligence and Digital Economy (SZ).